\documentclass{article}

\usepackage{microtype}
\usepackage{graphicx}
\usepackage{subfigure}
\usepackage{booktabs} 

\usepackage{hyperref}

\usepackage[accepted]{icml2021}

\usepackage{amsmath}
\usepackage{amsthm}
\usepackage{amssymb}
\usepackage{arydshln}
\usepackage[small,bf]{caption}
\usepackage{mathtools}
\usepackage{multirow}
\usepackage{enumitem}
\usepackage[normalem]{ulem}

\definecolor{darkred}{RGB}{162, 0, 0}
\newcommand{\darkred}[1]{\textbf{\textcolor{darkred}{#1}}}

\icmltitlerunning{Graph Contrastive Learning Automated}

\begin{document}

\twocolumn[
\icmltitle{Graph Contrastive Learning Automated}




\begin{icmlauthorlist}
\icmlauthor{Yuning You}{tamu}
\icmlauthor{Tianlong Chen}{ut}
\icmlauthor{Yang Shen}{tamu}
\icmlauthor{Zhangyang Wang}{ut}
\end{icmlauthorlist}

\icmlaffiliation{tamu}{Texas A\&M University}
\icmlaffiliation{ut}{The University of Texas at Austin}

\icmlcorrespondingauthor{Yang Shen}{yshen@tamu.edu}
\icmlcorrespondingauthor{Zhangyang Wang}{atlaswang@utexas.edu}

\icmlkeywords{Machine Learning, ICML}

\vskip 0.3in
]



\printAffiliationsAndNotice{}  

\begin{abstract}

Self-supervised learning on graph-structured data has drawn recent interest for learning generalizable, transferable and robust representations from unlabeled graphs. Among many, graph contrastive learning (GraphCL) has emerged with promising representation learning performance. Unfortunately, unlike its counterpart on image data, the effectiveness of GraphCL hinges on 
ad-hoc data augmentations, which have to be manually picked per dataset, by either rules of thumb or trial-and-errors, owing to the diverse nature of graph data. That significantly limits the more general applicability of GraphCL. Aiming to fill in this crucial gap, this paper proposes a unified bi-level optimization framework to \textbf{automatically}, \textbf{adaptively} and \textbf{dynamically} select data augmentations when performing GraphCL on specific graph data. The general framework, dubbed \textbf{JO}int \textbf{A}ugmentation \textbf{O}ptimization (\textbf{JOAO}), is instantiated as min-max optimization. The selections of augmentations made by JOAO are shown to be in general aligned with previous ``best practices" observed from handcrafted tuning: yet now being automated, more flexible and versatile. Moreover, we propose a new augmentation-aware projection head mechanism, which will route output features through different projection heads corresponding to different augmentations chosen at each training step. Extensive experiments demonstrate that JOAO performs on par with or sometimes better than the state-of-the-art competitors including GraphCL, on multiple graph datasets of various scales and types, yet without resorting to any laborious dataset-specific tuning on augmentation selection. We release the code at \url{https: //github.com/Shen-Lab/GraphCL_Automated}.

\end{abstract}

\section{Introduction}
Self-supervised learning on graph-structured data has raised significant interests recently \cite{hu2019strategies,you2020does,jin2020self,hu2020gpt,hwang2020self,manessi2020graph,zhu2020self,peng2020self,rong2020self,jin2021automated,roy2021node,huang2021hop}. Among many others, graph contrastive learning methods extend the contrastive learning idea \cite{he2020momentum,chen2020simple}, originally developed in the computer vision domain, to learn generalizable, transferable and robust representations from unlabeled graph data \cite{velivckovic2018deep,sun2019infograph, you2020graph,qiu2020gcc,hassani2020contrastive,zhu2020deep,zhu2020graph,chen2020distance,chen2020coad,ren2019heterogeneous,park2020unsupervised,peng2020graph,jin2021multi,wang2021learning}.
For comprehensive reviews of the topic, please refer to \cite{xie2021self,liu2021graph}.

Nevertheless, unlike images, graph datasets are abstractions of diverse nature (e.g. citation networks,  social networks, and biomedical networks at various levels ranging from molecules to healthcare systems \cite{li2021representation}). Such a unique diversity challenge was not fully addressed by prior graph self-supervised learning approaches \cite{hu2019strategies,you2020graph,you2020does}. For example, the state-of-the-art graph contrastive learning framework, GraphCL  \cite{you2020graph}, constructs specific contrastive views of graph data via hand-picking ad-hoc augmentations for every dataset \cite{you2020graph,zhao2020data,kong2020flag}. The choice of augmentation follows empirical rules of thumb, typically summarized from many trial-and-error experiments \textit{per dataset}. That seriously prohibits GraphCL and its variants from broader applicability, considering the tremendous heterogeneity of practical graph data. Moreover, even such trial-and-error selection of augmentations relies on a labeled validation set for downstream evaluation, which is not always available \cite{dwivedi2020benchmarking,hu2020open}.



\textbf{Contributions.} Given a new and unseen graph dataset, can our graph contrastive learning methods \text{automatically} select their data augmentation, avoiding ad-hoc choices or tedious tuning? This paper targets at overcoming this crucial, unique, and inherent hurdle. We propose \textit{joint augmentation optimization} (\textbf{JOAO}), a principled bi-level optimization framework that \textit{learns to select} data augmentations \textit{for the first time}. To highlight, the selection framework by JOAO is: (i) \textbf{automatic}, completely free of human labor of trial-and-error on augmentation choices; (ii) \textbf{adaptive}, generalizing smoothly to handling diverse graph data; and (iii) \textbf{dynamic}, allowing for augmentation types varying at different training stages. Compared to previous ad-hoc, per-dataset and pre-fixed augmentation selection, JOAO achieves an unprecedented degree of flexibility and ease of use. We summarize our contributions:
\begin{itemize}[leftmargin=*]
\vspace{-1em}
  \item Leveraging GraphCL \cite{you2020graph} as the baseline model, we introduce joint augmentation optimization (JOAO) as a plug-and-play framework. JOAO is the first to automate the augmentation selection when performing contrastive learning on specific graph data. It frees GraphCL from expensive trial-and-errors, or empirical ad-hoc rules, or any validation based on labeled data.\vspace{-0.5em}
  \item JOAO can be formulated as a unified bi-level optimization framework, and be instantiated as \textit{min-max optimization}. It takes inspirations from adversarial perturbations as data augmentations \cite{xie2020adversarial}, and can be solved by an alternating gradient-descent algorithm.\vspace{-0.5em} 
\item In accordance with diverse and dynamic augmentations enabled by JOAO, we design a new \textit{augmentation-aware} projection head for graph contrastive learning. The rationale is to avoid too many complicated augmentations distorting the original data distribution. The idea is to keep one nonlinear projection head per augmentation pair, and each time using the single head corresponding to the augmentation currently selected by JOAO.\vspace{-0.5em}
  \item Extensive experiments demonstrate that GraphCL with JOAO performs on par with or even sometimes better than state-of-the-art (SOTA) competitors, across multiple graph datasets of various types and scales, yet without resorting to tedious dataset-specific manual tuning or domain knowledge. We also show the augmentation selections made by JOAO are in general informed and often aligned with previous ``best practices".
\vspace{-0.5em}
\end{itemize}
We leave two additional remarks: \textbf{(1)} JOAO is designed to be flexible and versatile. Although this paper mainly demonstrates JOAO on GraphCL, they are not tied with each other. The general optimization formulation of JOAO allows it to be easily integrated with other graph contrastive learning frameworks too. \textbf{(2)} JOAO is designed for automating the tedious and ad-hoc augmentation selection. It intends to \textit{match} the state-of-the-art results achieved by exhaustive manual tuning, but not necessarily to surpass them all. To re-iterate, our aim is to scale up graph contrastive learning to numerous types and scales of graph data in the real world, via a hassle-free framework rather than tuning one by one.

\section{Preliminaries and Notations} \label{sec:preliminaries}

Graph neural networks (GNNs) have grown into powerful tools to model non-Euclidean graph-structured data arising from various fields  \cite{xu2018powerful,you2020l2,you2020cross,liu2020towards,zhang2020iterative}.
Let $G = \{V, E\}$\footnote{We use the sans-serif typeface to denote a random variable (e.g. $\mathsf{G}$). The same letter in the italic font (e.g. $G$) denotes a sample, and  the calligraphic font (e.g. $\mathcal{G}$) denotes the sample space.} denote an undirected graph in the space $\mathcal{G}$ with $V$ and $E$ being the set of nodes and edges, respectively, and   
$X_v \in \mathcal{R}^D$ for $v \in V$ being node features. A GNN is defined as the mapping $f: \mathcal{G} \rightarrow \mathcal{R}^{D'}$ that encodes a sample graph $G$ into an $D'$-dimensional vector.

Self-supervised learning on graphs is shown to learn more generalizable, transferable and robust graph representations, through exploiting vast unlabelled data \cite{jin2020self,hu2020gpt,hwang2020self,manessi2020graph,zhu2020self,peng2020self,rong2020self,jin2021automated,xie2021self,roy2021node,huang2021hop,li2021representation}. 
However, earlier self-supervised tasks often need to be carefully designed with domain knowledge \cite{you2020does,hu2019strategies} due to the intrinsic complicacy of graph datasets.

\begin{figure}[!htb]
\vspace{-0.5em}
\begin{center}
  \includegraphics[width=1\linewidth]{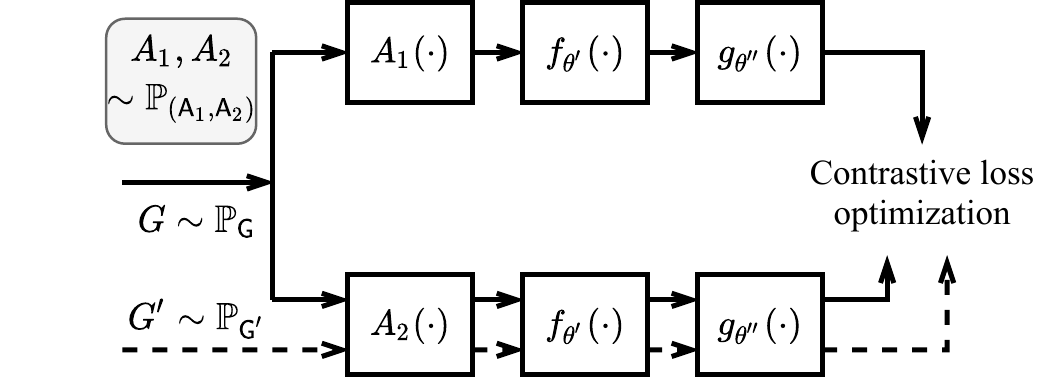}
\end{center}
\vspace{-0.7em}
  \caption{Overview of the GraphCL pipeline in \cite{you2020graph}.}
\label{fig:graphcl}
\vspace{-0.9em}
\end{figure}

Graph contrastive learning recently emerges as a promising direction \cite{velivckovic2018deep,sun2019infograph, qiu2020gcc,hassani2020contrastive,zhu2020deep,zhu2020graph,chen2020distance,chen2020coad,ren2019heterogeneous,park2020unsupervised,peng2020graph,jin2021multi,wang2021learning}. For example, the SOTA GraphCL framework \cite{you2020graph} enforces the perturbation invariance in GNNs through maximizing agreement between two augmented views of graphs: an overview is illustrated in Figure \ref{fig:graphcl}.

Specifically, we denote the input graph-structured sample $\mathsf{G}$ from certain empirical distribution $\mathbb{P}_{\mathsf{G}}$. Its samples two random augmentation operators $\mathsf{A}_1, \mathsf{A}_2$ from a given pool of augmentation types as $\mathcal{A} =  \{ \mathrm{NodeDrop}, \mathrm{Subgraph}, \mathrm{EdgePert}, \mathrm{AttrMask}, \mathrm{Identical} \}$ \cite{you2020graph} and $A \in \mathcal{A}: \mathcal{G} \rightarrow \mathcal{G}$.  
GraphCL \cite{you2020graph} optimizes the following loss:
\begin{align} \label{eq:graphcl}
    & \mathrm{min}_\theta \; \mathcal{L}(\mathsf{G}, \mathsf{A}_1, \mathsf{A}_2, \theta) &  \notag \\
    = & \, \mathrm{min}_\theta \; \Big\{ (- \mathbb{E}_{\mathbb{P}_{\mathsf{G}} \times
    \mathbb{P}_{(\mathsf{A}_1, \mathsf{A}_2)}
    } \mathrm{sim}( {\overbrace{\textstyle \mathsf{T}_{\theta, 1}(\mathsf{G}), \mathsf{T}_{\theta, 2}(\mathsf{G})}^{\mathclap{\text{Positive pairs}}} } ) \\
    & + \mathbb{E}_{\mathbb{P}_{\mathsf{G}} \times \mathbb{P}_{\mathsf{A}_1}} \mathrm{log}(\mathbb{E}_{\mathbb{P}_{\mathsf{G}'} \times \mathbb{P}_{\mathsf{A}_2}} \mathrm{exp}(\mathrm{sim}( {\underbrace{\mathsf{T}_{\theta, 1}(\mathsf{G}), \mathsf{T}_{\theta, 2}(\mathsf{G}')}_{\mathclap{\text{Negative pairs}}} })))\Big\}, \notag 
\end{align}
where $\mathsf{T}_{\theta, i} = \mathsf{A}_i \circ f_{\theta'} \circ g_{\theta''} \; (i=1,2)$ is parameterized by $\theta = \{ \theta', \theta'' \}$,
and $f_{\theta'}: \mathcal{G} \rightarrow \mathcal{R}^{D'}, g_{\theta''}: \mathcal{R}^{D'} \rightarrow \mathcal{R}^{D''}$ are the shared-weight GNN and projection head, respectively,
$\mathrm{sim}(u, v) = \frac{u^\mathsf{T}v}{\lVert u \rVert \lVert v \rVert}$ is the cosine similarity function,  $\mathbb{P}_{\mathsf{G}'} = \mathbb{P}_{\mathsf{G}}$ acts as the negative sampling distribution, and $\mathbb{P}_{\mathsf{A}_1}$ and $\mathbb{P}_{\mathsf{A}_2}$ are the marginal distributions. After the contrastive pre-training, the pre-trained $f_{\theta'^*}$ can be further leveraged for various downstream task fine-tuning.
 
In the current GraphCL framework, $(\mathsf{A}_1, \mathsf{A}_2)$ are selected by hand and pre-fixed for each dataset. In other words, $\mathbb{P}_{(\mathsf{A}_1, \mathsf{A}_2)}$ is a Dirac distribution with the only spike at the selected augmentation pair. Yet given new graph data, how to select $(\mathsf{A}_1, \mathsf{A}_2)$ relies on no more than loose heuristics. 






\section{Methodology}
\subsection{JOAO: The Unified Framework} \label{sec:joa}
One clear limitation in \eqref{eq:graphcl} is that one needs to pre-define the sampling distribution $\mathbb{P}_{(\mathsf{A}_1, \mathsf{A}_2)}$ based on prior rules, and only a Dirac distribution (i.e., only one pair for each dataset) was explored. Rather, we propose to dynamically and automatically learn to optimize  $\mathbb{P}_{(\mathsf{A}_1, \mathsf{A}_2)}$ when performing GraphCL \eqref{eq:graphcl}, via the following bi-level optimization framework:
\begin{align} \label{eq:auto_graphcl}
    & \mathrm{min}_\theta \quad \mathcal{L}(\mathsf{G}, \mathsf{A}_1, \mathsf{A}_2, \theta), \notag \\
    & \text{s.t.} \quad \mathbb{P}_{(\mathsf{A}_1, \mathsf{A}_2)} \in \mathrm{arg \, min}_{\mathbb{P}_{(\mathsf{A}_1', \mathsf{A}_2')}} \mathcal{D}(\mathsf{G}, \mathsf{A}_1', \mathsf{A}_2', \theta),
\end{align}
We refer to (\ref{eq:auto_graphcl}) as joint augmentation optimization (JOAO),
where the upper-level objective $\mathcal{L}$ is the same as the GraphCL objective (or the objective of any other graph contrastive learning approach), and the lower-level objective $\mathcal{D}$ optimizes the sampling distribution $\mathbb{P}_{(\mathsf{A}_1, \mathsf{A}_2)}$ jointly for augmentation-pair selections. Notice that JOAO \eqref{eq:auto_graphcl} only exploits the signals from the self-supervised training itself, without accessing downstream labeled data for evaluation.

\subsection{Instantiation of JOAO as Min-Max Optimization} \label{sec:minmax_graphcl}
Motivated from adversarial training \cite{wang2019towards,xie2020adversarial}, we follow the same philosophy to always exploit the most challenging data augmentation of the current loss, hence instantiating the general JOAO framework as a concrete min-max optimization form:
\begin{align} \label{eq:minmax_graphcl}
    & \mathrm{min}_\theta \quad \mathcal{L}(\mathsf{G}, \mathsf{A}_1, \mathsf{A}_2, \theta), \notag \\
    & \text{s.t.} \quad \mathbb{P}_{(\mathsf{A}_1, \mathsf{A}_2)} \in \mathrm{arg \, max}_{\mathbb{P}_{(\mathsf{A}_1', \mathsf{A}_2')}} \Big\{ \mathcal{L}(\mathsf{G}, \mathsf{A}_1', \mathsf{A}_2', \theta) \notag \\
    & \quad \quad \quad \quad \quad \quad - \frac{\gamma}{2} \, \mathrm{dist}(\mathrm{\mathbb{P}_{(\mathsf{A}_1', \mathsf{A}_2')}, \mathbb{P}_{\mathrm{prior}}}) \Big\},
\end{align}
where $\gamma \in \mathcal{R}_{\ge 0}$, $\mathbb{P}_{\mathrm{prior}}$ is the prior distribution on all possible augmentations, and $\mathrm{dist}: \mathcal{P} \times \mathcal{P} \rightarrow \mathcal{R}_{\ge 0}$  is a distance function between the sampling and the prior distribution ($\mathcal{P}$ is the probability simplex).
Thereby, JOAO's formulation aligns with the idea of model-based adversarial training \cite{robey2020model}, where adversarial training is known to boost generalization, robustness and transferability \cite{robey2020model,wang2019towards}.

In this work, we choose $\mathbb{P}_{\mathrm{prior}}$ as the uniform distribution to promote diversity in the selections, following a common principle of maximum entropy \cite{guiasu1985principle} in Bayesian learning.  No additional information is assumed about the dataset or the augmentation pool. In practice, it encourage\sout{d}s more diverse augmentation selections rather than collapsing to few.
Comparison between the formulations with and without the  prior is shown in Table S5 of Appendix E.
We use a squared Euclidean distance for $\mathrm{dist}(\cdot, \cdot)$.
Accordingly, we have  $\mathrm{dist}(\mathrm{\mathbb{P}_{(\mathsf{A}_1, \mathsf{A}_2)}, \mathbb{P}_{\mathrm{prior}}}) = \sum_{i=1}^{|\mathcal{A}|} \sum_{j=1}^{|\mathcal{A}|} ( p_{ij} - \frac{1}{|\mathcal{A}|^2} )^2 $ where the probability $p_{ij} = \mathrm{Prob}(\mathsf{A}_1 = A^i, \mathsf{A}_2 = A^j)$.



We will next present how to optimize \eqref{eq:minmax_graphcl}. Following \cite{wang2019towards}, we adopt the alternating gradient descent algorithm (AGD), alternating between upper-level minimization and lower-level maximization, as outlined in Algorithm \ref{alg:alternating_gradient_descent}. 

\begin{algorithm}[!htb]
   \caption{AGD for optimization \eqref{eq:minmax_graphcl}}
   \label{alg:alternating_gradient_descent}
\begin{algorithmic}
   \STATE {\bfseries Input:} initial parameter $\theta^{(0)}$, sampling distribution $\mathbb{P}_{(\mathsf{A}_1, \mathsf{A}_2)}^{(0)}$, optimization step $N$.
   \FOR{$n=1$ {\bfseries to} $N$}
   \STATE 1. Upper-level minimization: fix $\mathbb{P}_{(\mathsf{A}_1, \mathsf{A}_2)} = \mathbb{P}_{(\mathsf{A}_1, \mathsf{A}_2)}^{(n-1)}$, and call equation \eqref{eq:objective_minimization} to update $\theta^{(n)}$.
   \STATE 2. Lower-level maximization: fix $\theta = \theta^{(n)}$, and call equation \eqref{eq:design_maximization} to update $\mathbb{P}_{(\mathsf{A}_1, \mathsf{A}_2)}^{(n)}$.
   \ENDFOR
   \STATE {\bfseries Return:} Optimized parameter $\theta^{(N)}$.
\end{algorithmic}
\end{algorithm}

\textbf{Upper-level minimization.}
The upper-level minimization w.r.t. $\theta$ follows the conventional gradient descent procedure as in the GraphCL optimization \eqref{eq:graphcl} given the sampling distribution $\mathbb{P}_{(\mathsf{A}_1, \mathsf{A}_2)}$, represented as:
\begin{equation} \label{eq:objective_minimization}
    \theta^{(n)} = \theta^{(n-1)} - \alpha' \triangledown_\theta \mathcal{L}(\mathsf{G}, \mathsf{A}_1, \mathsf{A}_2, \theta),
\end{equation}
where $\alpha' \in \mathcal{R}_{> 0}$ is the learning rate.


\textbf{Lower-level maximization.}
Since it is not intuitive to directly calculate the gradient of the lower-level objective w.r.t. $\mathbb{P}_{(\mathsf{A}_1, \mathsf{A}_2)}$, we first rewrite the contrastive loss in \eqref{eq:graphcl} as:
\begin{align} \label{eq:graphcl_loss}
    & \mathcal{L}(\mathsf{G}, \mathsf{A}_1, \mathsf{A}_2, \theta) = \sum_{i=1}^{|\mathcal{A}|} \sum_{j=1}^{|\mathcal{A}|} \;
    {\overbrace{\textstyle p_{ij}}^{\mathclap{\text{Targeted}}} }
    \Big\{ - \mathbb{E}_{\mathbb{P}_{\mathsf{G}}} \mathrm{sim}(T_\theta^i(\mathsf{G}), T_\theta^j(\mathsf{G})) \notag \\
    & \quad \quad + \mathbb{E}_{\mathbb{P}_{\mathsf{G}}} \mathrm{log}(\sum_{j'=1}^{|\mathcal{A}|} \;\;
    {\underbrace{\textstyle p_{j'}}_{\mathclap{\text{Undesired}}} }
    \mathbb{E}_{\mathbb{P}_{\mathsf{G}'}} \mathrm{exp}(\mathrm{sim}(T_\theta^i(\mathsf{G}), T_\theta^{j'}(\mathsf{G}')))) \Big\},
\end{align}
where $T_\theta^i = A^i \circ f_{\theta'} \circ g_{\theta''},\;(i=1, ..., 5)$, and the marginal probabilities $p_{j'} = p_j = \mathrm{Prob}(\mathsf{A}_2 = A^j)$.
In the equation \eqref{eq:graphcl_loss}, we expand the expectation on augmentations $\mathsf{A}_1, \mathsf{A}_2$ into the form of weighted summation related to $p_{ij}$ in order to calculate the gradient.
However, within the expectation on $\mathsf{G}$ of the negative pair term there is the marginal probabilities $p_{j'}$ entangled,
and therefore we make the following numerical approximation for the lower bound of the negative pair term to disentangle $p_{ij}$ in the equation \eqref{eq:graphcl_loss}:
\begin{align} \label{eq:approximation}
    & \mathbb{E}_{\mathbb{P}_{\mathsf{G}} \times \mathbb{P}_{\mathsf{A}_1}} \mathrm{log}(\mathbb{E}_{\mathbb{P}_{\mathsf{G}'} \times \mathbb{P}_{\mathsf{A}_2}} \mathrm{exp}(\mathrm{sim}( \mathsf{T}_{\theta, 1}(\mathsf{G}), \mathsf{T}_{\theta, 2}(\mathsf{G}') ))) \notag \\
    \ge & \, \mathbb{E}_{\mathbb{P}_{\mathsf{G}} \times \mathbb{P}_{\mathsf{A}_1} \times \mathbb{P}_{\mathsf{A}_2}} \mathrm{log}(\mathbb{E}_{\mathbb{P}_{\mathsf{G}'}} \mathrm{exp}(\mathrm{sim}( \mathsf{T}_{\theta, 1}(\mathsf{G}), \mathsf{T}_{\theta, 2}(\mathsf{G}') ))) \notag \\
    \approx & \, \mathbb{E}_{\mathbb{P}_{\mathsf{G}} \times \mathbb{P}_{(\mathsf{A}_1, \mathsf{A}_1)}} \mathrm{log}(\mathbb{E}_{\mathbb{P}_{\mathsf{G}'}} \mathrm{exp}(\mathrm{sim}( \mathsf{T}_{\theta, 1}(\mathsf{G}), \mathsf{T}_{\theta, 2}(\mathsf{G}') ))),
\end{align}
where the first inequality comes from Jensen's inequality, and the second approximation is numerical.
It results in the approximated contrastive loss:
\begin{align} \label{eq:graphcl_loss_approximation}
    & \mathcal{L}(\mathsf{G}, \mathsf{A}_1, \mathsf{A}_2, \theta) \approx \sum_{i=1}^{|\mathcal{A}|} \sum_{j=1}^{|\mathcal{A}|} \;
    {\overbrace{\textstyle p_{ij}}^{\mathclap{\text{Targeted}}} }
    \ell(\mathsf{G}, A^i, A^j, \theta) \notag \\
    = & \sum_{i=1}^{|\mathcal{A}|} \sum_{j=1}^{|\mathcal{A}|} p_{ij} \Big\{ - \mathbb{E}_{\mathbb{P}_{\mathsf{G}}}  \mathrm{sim}(T_\theta^i(\mathsf{G}), T_\theta^j(\mathsf{G})) \notag \\
    & + \mathbb{E}_{\mathbb{P}_{\mathsf{G}}} \mathrm{log}(\mathbb{E}_{\mathbb{P}_{\mathsf{G}'}} \mathrm{exp}(\mathrm{sim}(T_\theta^i(\mathsf{G}), T_\theta^j(\mathsf{G}')))) \Big\}.
\end{align}
Through approximating the contrastive loss, the lower-level maximization in the optimization \eqref{eq:minmax_graphcl} is rewritten as:
\begin{align} \label{eq:design_maximization_approximation}
    & \mathbb{P}_{(\mathsf{A}_1, \mathsf{A}_2)} \in \mathrm{arg \, max}_{\boldsymbol{p} \in \mathcal{P}, \boldsymbol{p} = [p_{ij}], i, j = 1, ..., |\mathcal{A}|} \{ \psi(\boldsymbol{p}) \}, \notag \\
    & \psi(\boldsymbol{p}) = \sum_{i=1}^{|\mathcal{A}|} \sum_{j=1}^{|\mathcal{A}|} p_{ij} \ell(\mathsf{G}, A^i, A^j, \theta) - \frac{\gamma}{2} \sum_{i=1}^{|\mathcal{A}|} \sum_{j=1}^{|\mathcal{A}|} ( p_{ij} - \frac{1}{|\mathcal{A}|^2} )^2,
\end{align}
\begin{figure}[t]
\begin{center}
  \includegraphics[width=1\linewidth]{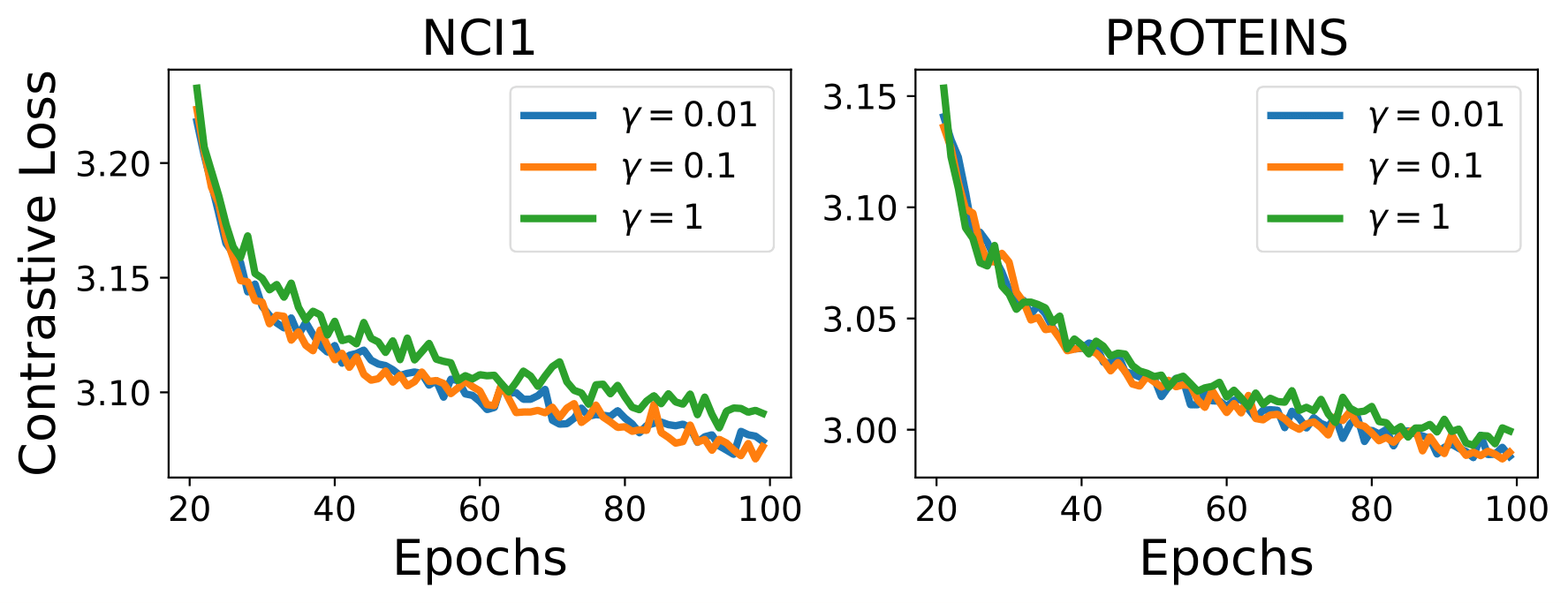}
\end{center}
\vspace{-4mm}
\caption{Empirical training curves of AGD in JOAO on datasets NCI1 and PROTEINS with different $\gamma$ values.}
 \vspace{-1em}
\label{fig:convergence}
\end{figure}
\noindent where $\psi(\boldsymbol{p})$ is a strongly-concave function w.r.t. $\boldsymbol{p}$ in the probability simplex $\mathcal{P}$.
Thus, a projected gradient descent \cite{wang2019towards,boyd2004convex} is performed to update the sampling distribution $\mathbb{P}_{(\mathsf{A}_1, \mathsf{A}_2)}$ for selecting augmentation pairs, expressed as:
\begin{equation} \label{eq:design_maximization}
    \boldsymbol{b} = \boldsymbol{p}^{(n-1)} + \alpha'' \triangledown_{\boldsymbol{p}} \psi(\boldsymbol{p}^{(n-1)}), \boldsymbol{p}^{(n)} = (\boldsymbol{b} - \mu \boldsymbol{1})_+,
\end{equation}
where $\alpha'' \in \mathcal{R}_{> 0}$ is the learning rate, $\mu$ is the root of the equation $\boldsymbol{1}^{\mathsf{T}} (\boldsymbol{b} - \mu \boldsymbol{1}) = 1$, and $(\cdot)_+$ is the element-wise non-negative operator. $\mu$ can be efficiently found via the bi-jection method \cite{wang2019towards,boyd2004convex}.


Even though an optimizer with theoretical guarantee of convergence for non-convex non-concave min-max problems remains an open challenge, we acknowledge that AGD is an approximation of solving the bi-level optimization \eqref{eq:minmax_graphcl} precisely, which typically costs Bayesian optimization \cite{srinivas2010gaussian,snoek2012practical}, automatic differentiation \cite{luketina2016scalable,franceschi2017forward,baydin2017online,shaban2019truncated}, or first-order techniques based on some inner-loop approximated solution \cite{maclaurin2015gradient,pedregosa2016hyperparameter,gould2016differentiating}. As most of them suffer from high time or space complexity, AGD was adopted as an approximated heuristic mainly for saving computational overhead. It showed some level of empirical convergence as seen in Figure \ref{fig:convergence}.



\subsubsection{Sanity Check: JOAO Recovers Augmentation-Pairs Aligned with Previous ``Best Practices"} \label{sec:joa_correlation}
How reasonable are the JOAO-selected augmentation pairs per dataset? This section pass JOAO through a sanity check, by comparing its selections with the previous trial-and-error findings by manually and exhaustively combining different augmentations (using downstream labels for validation) \cite{you2020graph}. 

\begin{table}[!htb]
\vspace{-0.5em}
 \caption{Datasets statistics.}
 \vspace{-0.5em}
 \label{tab:statistics}
 \centering
 \resizebox{0.48\textwidth}{!}{
 \begin{tabular}{c | c | c | c | c } 
  \hline
  \hline
  Datasets & Category & Graph Num. & Avg. Node & Avg. Degree \\
  \hline
  \hline
  NCI1 & Biochemical Molecules & 4110 & 29.87 & 1.08 \\
  PROTEINS & Biochemical Molecules & 1113 & 39.06 & 1.86 \\
  \hline
  COLLAB & Social Networks & 5000 & 74.49 & 32.99 \\
  RDT-B & Social Networks & 2000 & 429.63 & 1.15 \\
  \hline
  \hline
 \end{tabular}}
 \vspace{-0.5em}
\end{table}




\begin{figure*}[t]
\begin{center}
  \includegraphics[width=0.95\linewidth]{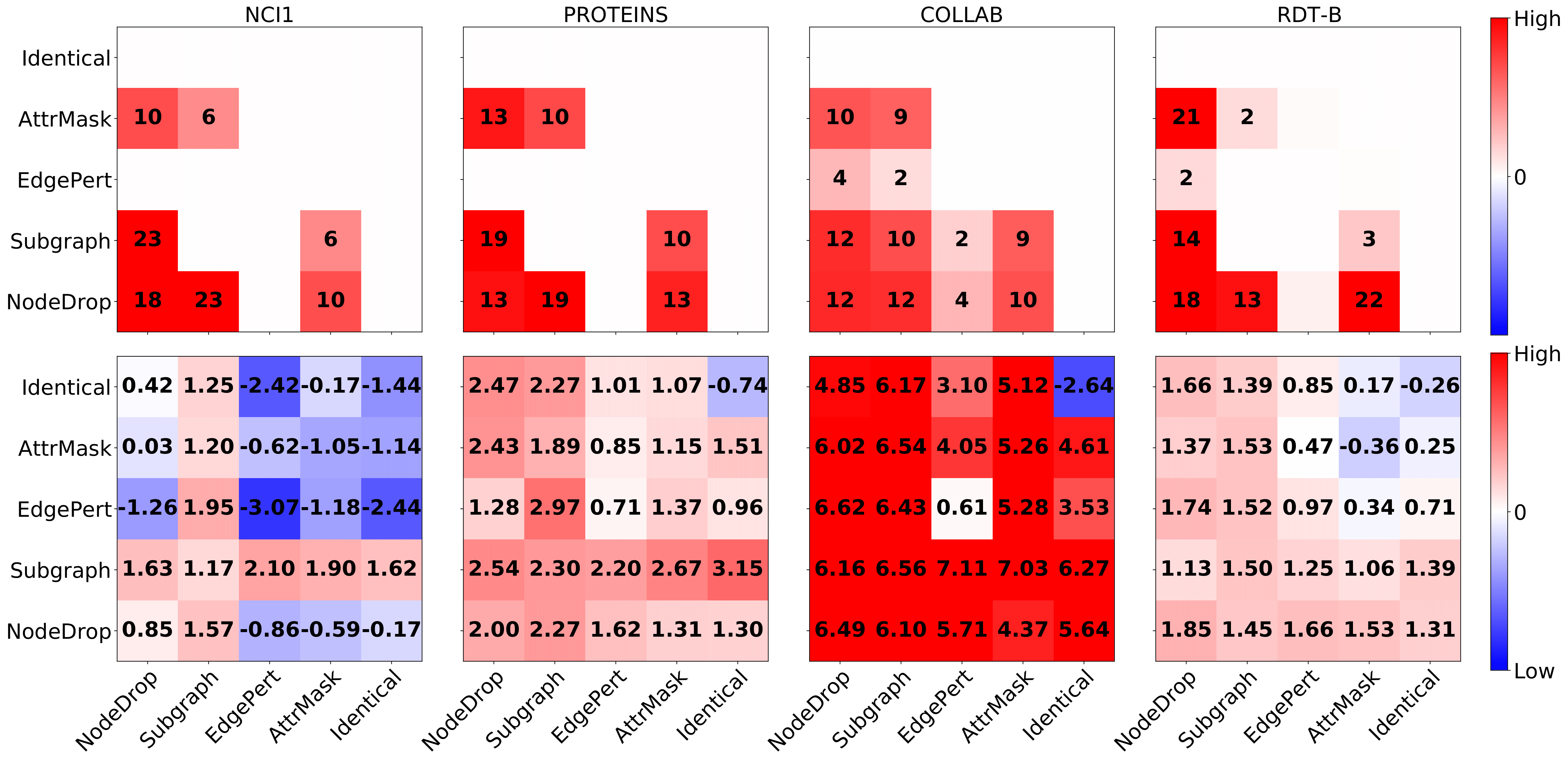}
\end{center}
  \caption{\textbf{Top row}: sampling distributions (\%, defined as the percentage of this specific augmentation pair being selected during the entire training process) for augmentation pairs selected by JOAO on four different datasets (NCI1, PROTEINS, COLLAB, and RDT-B ). \textbf{Bottom row}: GraphCL performance gains (classification accuracy \%, see \cite{you2020graph} for the detailed setting) when exhaustively trying every possible augmentation pair. \textbf{Note that} the percentage numbers in the first and second rows have different meanings and are not apple-to-apple comparable; however, the overall alignments between the two rows' trends and high-value locations indicate that, if an augmentation pair was manually verified to yield better GraphCL results, it is also more likely to be selected by JOAO. Warmer (colder) colors indicate higher (lower) values, and white marks 0.
}
\vspace{-0.5em}
\label{fig:prob}
\end{figure*}

To examine such alignment, we visualize in the top row of Figure \ref{fig:prob} the JOAO-optimized sampling distributions $\mathbb{P}_{(\mathsf{A}_1, \mathsf{A}_2)}$, and in the bottom row the GraphCL's manual trial-and-error results over various augmentation pairs, for four different datasets (data statistics in Table \ref{tab:statistics}). Please refer to the caption on Figure \ref{fig:prob} how to interpret the percentage numbers in the top and bottom rows respectively. Overall, we observe a decent extent of alignments between the two rows’ trends and especially the high-value locations, indicating that: if an augmentation pair was manually verified to yield better GraphCL results, it is also more likely to be selected by JOAO. More specifically we can see: (1) augmentation pairs containing EdgePert and AttrMask are more likely to be selected for  biochemical molecules and denser graphs, respectively; (2) NodeDrop and Subgraph are generally adopted on all four datasets; and (3) the augmentation pairs of two identy transformations are completely abandoned by JOAO, and those pairs with more diverse transformation types are more desired. All those observations are well aligned with the ``rules of thumb" summarized in \cite{you2020graph}. More discussions are provided in Appendix D.

Therefore, the selections of augmentations made by JOAO are shown to be generally consistent with previous ``best practices" observed from manual tuning – yet now being fully automated, flexible, versatile. It is also achieved without using any downstream task label, while \cite{you2020graph} would hinge on a labeled set to compare two augmentations by their downstream performance.

\subsection{Augmentation-Aware Multi-Projection Heads: Addressing A New Challenge from JOAO} \label{sec:joap}

JOAO conveys the blessing of diverse and dynamic augmentations that are selected automatically during each GraphCL training, which may yield more robust and invariant features. However, that blessing could also bring up a new challenge: compared to one fixed augmentation pair throughout training, those varying and more aggressive augmentations can distort the training distribution more \cite{lee2020self,jun2020distribution}. Even mild augmentations, such as adding/dropping nodes or edges, could result in graphs very unlikely under the original distribution. Models trained with these augmentations may fit the original distribution poorly.


To address this challenge arising from using JOAO, we introduce multiple projection heads and an augmentation-aware selection scheme into GraphCL, as glimpsed in Figure \ref{fig:graphcl_proj} (see Figure S2 in Appendix D for a schematic diagram).
Specifically, we construct $|\mathcal{A}|$ projection heads
each of which corresponds to one augmentation type ($|\mathcal{A}|$ denotes the cardinality of the augmentation pool).
Then during training, once an augmentation is sampled, it will only go through and update its corresponding projection head.
The main idea is to explicitly disentangle the distorted feature distributions caused by various augmentation pairs, and each time we only use the one head corresponding to the augmentation currently selected by JOAO.
\begin{figure}[!htb]
\begin{center}
\vspace{-0.5em}
  \includegraphics[width=1\linewidth]{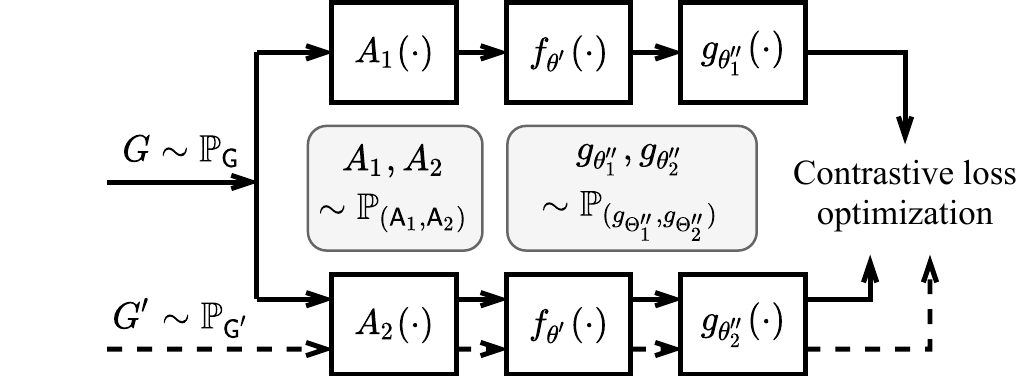}
\end{center}
\vspace{-0.5em}
  \caption{An overview of GraphCL with multiple augmentation-aware projection heads where $\mathbb{P}_{(g_{{\Theta''_1}}, g_{{\Theta''_2}})} = \mathbb{P}_{(\mathsf{A}_1, \mathsf{A}_2)}$.}
  \vspace{-0.5em}
\label{fig:graphcl_proj}
\end{figure}

In mathematical forms, we route the output features from $f$ through the projection head sampled from $\mathbb{P}_{(g_{{\Theta''_1}}, g_{{\Theta''_2}})}$ at each training step, where $\mathbb{P}_{(g_{{\Theta''_1}}, g_{{\Theta''_2}})} = \mathbb{P}_{(\mathsf{A}_1, \mathsf{A}_2)}$, and $\Theta''_1, \Theta''_2$ denote the head parameters, resulting in $\mathsf{T}_{\theta, i} = \mathsf{A}_i \circ f_{\theta'} \circ g_{\Theta''_i}, \,(i=1,2)$. 
Denoting $\mathcal{L}_{\text{v2}}(\mathsf{G}, \mathsf{A}_1, \mathsf{A}_2, \theta', \Theta''_1, \Theta''_2) = \mathbb{E}_{\mathbb{P}_{(g_{{\Theta''_1}}, g_{{\Theta''_2}})}} \mathcal{L}(\mathsf{G}, \mathsf{A}_1, \mathsf{A}_2,$ $\{\theta', (\Theta''_1, \Theta''_2)\})$, we could then integrate the augmentation-aware projection head mechanism into the JOAO framework, referred to as \textbf{JOAOv2}:
\begin{align} \label{eq:minmax_graphcl_joint}
    & \mathrm{min}_\theta \quad \mathcal{L}_{\text{v2}}(\mathsf{G}, \mathsf{A}_1, \mathsf{A}_2, \theta', \Theta''_1, \Theta''_2), \notag \\
    & \text{s.t.} \quad \mathbb{P}_{(\mathsf{A}_1, \mathsf{A}_2)} \in \mathrm{arg \, max}_{\mathbb{P}_{(\mathsf{A}_1', \mathsf{A}_2')}} \Big\{ \mathcal{L}_{\text{v2}}(\mathsf{G}, \mathsf{A}_1, \mathsf{A}_2, \theta', \Theta''_1, \Theta''_2) \notag \\
    & \quad \quad \quad \quad \quad \quad - \frac{\gamma}{2} \, \mathrm{dist}(\mathrm{\mathbb{P}_{(\mathsf{A}_1', \mathsf{A}_2')}, \mathbb{P}_{\mathrm{prior}}}) \Big\}, \notag \\
    & \quad \quad \, \mathbb{P}_{(g_{{\Theta''_1}}, g_{{\Theta''_2}})} = \mathbb{P}_{(\mathsf{A}_1, \mathsf{A}_2)}.
\end{align}
Algorithm \ref{alg:alternating_gradient_descent} could be easily adapted to solve \eqref{eq:minmax_graphcl_joint} (see Algorithm S1 of Appendix A).

Our preliminary experiments in Table \ref{tab:aug_ratio} show that, without bells and whistles, augmentation-aware projection heads improve the performance upon JOAO under different augmentation strengths. That  aligns  with the observations in \cite{lee2020self,jun2020distribution}, showing that disentangling augmented and original feature distributions could have the model benefit more from stronger augmentations. 
 \vspace{-0.5em}
\begin{table}[ht] 
 \caption{Experiments with JOAO and JOAOv2 without explicit hyper-parameter tuning under different augmentation strengths on NCI1 and PROTEINS. A.S. is short for augmentation strength.}
 \label{tab:aug_ratio}
 \centering
 \resizebox{0.35\textwidth}{!}{
 \begin{tabular}{c | c | c c} 
  \hline
  \hline
  Datasets & A.S. & JOAO & JOAOv2 \\
  \hline
  \hline
  \multirow{2}{*}{NCI1} & 0.2 & 61.77$\pm$1.61 & 62.52$\pm$1.16 \\
   & 0.25 & 60.95$\pm$0.55 & 61.67$\pm$0.72 \\
   \hline
   \multirow{2}{*}{PROTEINS} & 0.2 & 71.45$\pm$0.89 & 71.66$\pm$1.10 \\
   & 0.25 & 71.61$\pm$1.65 & 73.01$\pm$1.02 \\
  \hline
  \hline
 \end{tabular}}
\end{table} \vspace{-0.5em}

We also plot the learned $\mathbb{P}_{(\mathsf{A}_1, \mathsf{A}_2)}$ in Figure S1 of Appendix D, where we can observe an ever stronger alignment than presented in Figure \ref{fig:prob}.

\section{Experiments}  
In this section, we evaluate our proposed methods, JOAO and JOAOv2, against state-of-the-art (SOTA) \textit{competitors} including  self-supervised approaches heuristically designed with domain knowledge, and graph contrastive learning (GraphCL) with  pre-defined rules for augmentation selection,
using the \textit{scenarios} of
datasets originated from diverse sources,
and datasets on specific bioinformatics domains.
An summary of main results can be found in Table \ref{tab:summary}.
\begin{table}[ht]
 \caption{Summary of JOAO performance.}
 \label{tab:summary}
 \centering
 \resizebox{0.45\textwidth}{!}{
 \begin{tabular}{c | c | c } 
  \hline
  \hline
   & v.s. GraphCL & v.s. Heuristic methods \\
  \hline
  \hline
  Across diverse fields & Comparable & Better \\
  \hline
  On specific domains & Better & Worse \\
  \hline
  \hline
 \end{tabular}}
\end{table}

\subsection{Datasets and Experiment Settings}
\textbf{Datasets.}
We use datasets of diverse nature from the benchmark TUDataset \cite{morris2020tudataset}, including graph data for small molecules \& proteins \cite{riesen2008iam,dobson2003distinguishing}, computer vision \cite{nene1996columbia} and various relation networks \cite{yanardag2015deep,rozemberczki2020api} of diverse statistics (see Table S1 of Appendix B), under semi-supervised and unsupervised learning.
Additionally we gather domain-specific bioinformatics datasets from the benchmark \cite{hu2019strategies} of relatively similar statistics (see Table S2 of Appendix B), under transfer-learning tasks for predicting molecules' chemical property or proteins' biological function.
Lastly we take two large-scale benchmark datasets, ogbg-ppa \& ogbg-code from Open Graph Benchmark (OGB) \cite{hu2020open} (see Table S3 of Appendix B for statistics) to evaluate scalability under semi-supervised learning.  

\textbf{Learning protocols.}
Learning experiments are performed in three settings, following the same protocols as in SOTA work. (1) In semi-supervised learning \cite{you2020graph} on datasets without the explicit training/validation/test split, we perform pre-training with all data and did finetuning \& evaluation with $K$ folds where $K = \frac{1}{\text{label rate}}$; 
and on datasets with the train/validation/test split, we only perform pre-training with the training data, finetuning on the partial training data and evaluation on the validation/test sets.
(2) In unsupervised representation learning \cite{sun2019infograph}, we pre-train using the whole dataset to learn graph embeddings and feed them into a downstream SVM classifier with 10-fold cross-validation.
(3) In transfer learning \cite{hu2019strategies}, we pre-train on a larger dataset then finetune and evaluate on smaller datasets of the same category 
using the given training/validation/test split.

\textbf{GNN architectures \& augmentations.}
We adopt the same GNN architectures with default hyper-parameters as in the SOTA methods under individual experiment settings.
Specifically, (1) in semi-supervised learning, ResGCN \cite{chen2019powerful} is used with 5 layers and 128 hidden dimensions, (2) in unsupervised representation learning, GIN \cite{xu2018powerful} is used with 3 layers and 32 hidden dimensions, and (3) in transfer learning and on large-scale OGB datasets, GIN is used with 5 layers and 300 hidden dimensions.
Plus, we adopt the same graph data augmentations as in GraphCL \cite{you2020graph} with the default augmentation strength 0.2.
We tune the hyper-parameter $\gamma$ controlling the trade-off in the optimization \eqref{eq:minmax_graphcl} in the range of \{0.01, 0.1, 1\}.
\begin{table*}[t]
\begin{center}
\caption{
Semi-supervised learning on TUDataset.
Shown in \darkred{red} are the best accuracy (\%) and those within the standard deviation of the best accuracy or the best average ranks.  - indicates that label rate is too low for a given dataset size.
L.R. and A.R. are short for label rate and average rank, respectively.
The compared results except those for ContextPred are as published under the same experiment setting.
}
\label{tab:semi-supervised}
\resizebox{0.85\textwidth}{!}{\begin{tabular}{c | c | c c c | c c c c | c}
    \hline
    \hline
    L.R. & Methods & NCI1 & PROTEINS & DD & COLLAB & RDT-B & RDT-M5K & GITHUB & A.R.$\downarrow$ \\
    \hline
    \hline
    1\% & No pre-train. & 60.72$\pm$0.45 & - & - & 57.46$\pm$0.25 & - & - & 54.25$\pm$0.22 & 7.6 \\
    & Augmentations & 60.49$\pm$0.46 & - & - & 58.40$\pm$0.97 & - & - & 56.36$\pm$0.42 & 6.6 \\
    \cdashline{2-10}
    & GAE & 61.63$\pm$0.84 & - & - & 63.20$\pm$0.67 & - & - & 59.44$\pm$0.44 & 4.0 \\
    & Infomax & \darkred{62.72}$\pm$0.65 & - & - & 61.70$\pm$0.77 & - & - & 58.99$\pm$0.50 & 3.3 \\
    & ContextPred & 61.21$\pm$0.77 & - & - & 57.60$\pm$2.07 & - & - & 56.20$\pm$0.49 & 6.6 \\
    & GraphCL & \darkred{62.55}$\pm$0.86 & - & - & \darkred{64.57}$\pm$1.15 & - & - & 58.56$\pm$0.59 & \darkred{2.6} \\
    \cdashline{2-10}
    & JOAO & 61.97$\pm$0.72 & - & - & \darkred{63.71}$\pm$0.84 & - & - & 60.35$\pm$0.24 & 3.0 \\
    & JOAOv2 & \darkred{62.52}$\pm$1.16 & - & - & \darkred{64.51}$\pm$2.21 & - & - & \darkred{61.05}$\pm$0.31 & \darkred{2.0} \\
    \hline
    10\% & No pre-train. & 73.72$\pm$0.24 & 70.40$\pm$1.54 & 73.56$\pm$0.41 & 73.71$\pm$0.27 & 86.63$\pm$0.27 & 51.33$\pm$0.44 & 60.87$\pm$0.17 & 7.0 \\
    & Augmentations & 73.59$\pm$0.32 & 70.29$\pm$0.64 & 74.30$\pm$0.81 & 74.19$\pm$0.13 & 87.74$\pm$0.39 & 52.01$\pm$0.20 & 60.91$\pm$0.32 & 6.2 \\
    \cdashline{2-10}
    & GAE & 74.36$\pm$0.24 & 70.51$\pm$0.17 & 74.54$\pm$0.68 & 75.09$\pm$0.19 & 87.69$\pm$0.40 & \darkred{53.58}$\pm$0.13 & 63.89$\pm$0.52 & 4.5 \\
    & Infomax & \darkred{74.86}$\pm$0.26 & 72.27$\pm$0.40 & \darkred{75.78}$\pm$0.34 & 73.76$\pm$0.29 & 88.66$\pm$0.95 & \darkred{53.61}$\pm$0.31 & 65.21$\pm$0.88 & 3.0 \\
    & ContextPred & 73.00$\pm$0.30 & 70.23$\pm$0.63 & 74.66$\pm$0.51 & 73.69$\pm$0.37 & 84.76$\pm$0.52 & 51.23$\pm$0.84 & 62.35$\pm$0.73 & 7.2 \\
    & GraphCL & \darkred{74.63}$\pm$0.25 & \darkred{74.17}$\pm$0.34 & \darkred{76.17}$\pm$1.37 & 74.23$\pm$0.21 & \darkred{89.11}$\pm$0.19 & 52.55$\pm$0.45 & 65.81$\pm$0.79 & \darkred{2.4} \\
    \cdashline{2-10}
    & JOAO & 74.48$\pm$0.27 & 72.13$\pm$0.92 & \darkred{75.69}$\pm$0.67 & \darkred{75.30}$\pm$0.32 & 88.14$\pm$0.25 & 52.83$\pm$0.54 & 65.00$\pm$0.30 & 3.5 \\
    & JOAOv2 & \darkred{74.86}$\pm$0.39 & 73.31$\pm$0.48 & \darkred{75.81}$\pm$0.73 & \darkred{75.53}$\pm$0.18 & 88.79$\pm$0.65 & 52.71$\pm$0.28 & \darkred{66.66}$\pm$0.60 & \darkred{1.8} \\
    \hline
    \hline
\end{tabular}}
\end{center}
\end{table*}

\begin{table*}[t]
\begin{center}
\caption{Unsupervised representation learning on TUDataset.
\darkred{Red} numbers indicate the top-3 accuracy (\%) or the top-2 average ranks.
The compared results are from the published papers, and
- indicates that results were not available in published papers. For MVGRL we report the numbers with the NT-Xent loss to be comparable with GraphCL.
}
\label{tab:unsupervised}
\resizebox{0.9\textwidth}{!}{
\begin{tabular}{c | c c c c | c c c c | c }
    \hline
    \hline
    Methods & NCI1 & PROTEINS & DD & MUTAG & COLLAB & RDT-B & RDT-M5K & IMDB-B & A.R.$\downarrow$ \\
    \hline
    \hline
    GL & - & - & - & 81.66$\pm$2.11 & - & 77.34$\pm$0.18 & 41.01$\pm$0.17 & 65.87$\pm$0.98 & 7.4 \\
    WL & \darkred{80.01}$\pm$0.50 & 72.92$\pm$0.56 & - & 80.72$\pm$3.00 & - & 68.82$\pm$0.41 & 46.06$\pm$0.21 & \darkred{72.30}$\pm$3.44 & 5.7 \\
    DGK & \darkred{80.31}$\pm$0.46 & 73.30$\pm$0.82 & - & 87.44$\pm$2.72 & - & 78.04$\pm$0.39 & 41.27$\pm$0.18 & 66.96$\pm$0.56 & 4.9 \\
    \hline
    node2vec & 54.89$\pm$1.61 & 57.49$\pm$3.57 & - & 72.63$\pm$10.20 & - & - & - & - & 8.6 \\
    sub2vec & 52.84$\pm$1.47 & 53.03$\pm$5.55 & - & 61.05$\pm$15.80 & - & 71.48$\pm$0.41 & 36.68$\pm$0.42 & 55.26$\pm$1.54 & 9.5 \\
    graph2vec & 73.22$\pm$1.81 & 73.30$\pm$2.05 & - & 83.15$\pm$9.25 & - & 75.78$\pm$1.03 & 47.86$\pm$0.26 & 71.10$\pm$0.54 & 5.7 \\
    MVGRL & - & - & - & 75.40$\pm$7.80 & - & 82.00$\pm$1.10 & - & 63.60$\pm$4.20 & 7.2 \\
    InfoGraph & 76.20$\pm$1.06 & \darkred{74.44}$\pm$0.31 & 72.85$\pm$1.78 & \darkred{89.01}$\pm$1.13 & \darkred{70.65}$\pm$1.13 & 82.50$\pm$1.42 & 53.46$\pm$1.03 & \darkred{73.03}$\pm$0.87 & 3.0 \\
    GraphCL & 77.87$\pm$0.41 & \darkred{74.39}$\pm$0.45 & \darkred{78.62}$\pm$0.40 & 86.80$\pm$1.34 & \darkred{71.36}$\pm$1.15 & \darkred{89.53}$\pm$0.84 & \darkred{55.99}$\pm$0.28 & \darkred{71.14}$\pm$0.44 & \darkred{2.6} \\
    \hline
    JOAO & \darkred{78.07}$\pm$0.47 & \darkred{74.55}$\pm$0.41 & \darkred{77.32}$\pm$0.54 & \darkred{87.35}$\pm$1.02 & \darkred{69.50}$\pm$0.36 & \darkred{85.29}$\pm$1.35 & \darkred{55.74}$\pm$0.63 & 70.21$\pm$3.08 & 3.3 \\
    JOAOv2 & 78.36$\pm$0.53 & 74.07$\pm$1.10 & \darkred{77.40}$\pm$1.15 & \darkred{87.67}$\pm$0.79 & 69.33$\pm$0.34 & \darkred{86.42}$\pm$1.45 & \darkred{56.03}$\pm$0.27 & 70.83$\pm$0.25 & \darkred{2.8} \\
    \hline
    \hline
\end{tabular}}
\end{center}
\end{table*}

\subsection{Compared Algorithms.} \label{sec:compared_algorithms}
\textbf{Training from scratch (with augmentations) and graph kernels.}
The na\"ive baseline training from the random initialization (with same augmentations as in GraphCL \cite{you2020graph}) is compared,
as well as SOTA graph kernel methods including GL \cite{shervashidze2009efficient}, WL \cite{shervashidze2011weisfeiler} and DGK \cite{yanardag2015deep}.

\textbf{Heuristic self-supervised methods.}
Heuristic self-supervised methods are designed based on certain domain knowledge, which work well when such knowledge is available and benefits downstream tasks.
The compared ones include:
(1) edge-based reconstruction including GAE \cite{kipf2016variational}, node2vec \cite{grover2016node2vec} and EdgePred \cite{hu2019strategies},
(2) vertex feature masking \& recover, namely AttrMasking \cite{hu2019strategies},
(3) sub-structure information preserving such as sub2vec \cite{adhikari2018sub2vec}, graph2vec \cite{narayanan2017graph2vec} and ContextPred \cite{hu2019strategies},
and (4) global-local representation consistency such as Infomax \cite{velivckovic2018deep} \& InfoGraph \cite{sun2019infograph}.
We adopt the default hyper-parameters published for these methods.

\textbf{GraphCL with pre-fixed augmentation sampling rules.}
For constructing the sampling pool of augmentations, we follow the same rule as in \cite{you2020graph} that uses 
(1) NodeDrop and Subgraph for biochemical molecules,
(2) all for dense relation networks,
and (3) all except AttrMask for sparse relation networks.
The exact augmentations for each dataset are shown in Table S4 of Appendix C.
\begin{table*}[t]
\begin{center}
\caption{Transfer learning on bioinformatics datasets.
\darkred{Red} numbers indicate the top-3 performances (AUC of ROC in \%).
Results for SOTA methods are as published.
}
\label{tab:transfer}
\resizebox{0.95\textwidth}{!}{
\begin{tabular}{c | c c c  c c c c c | c | c }
    \hline
    \hline
    Methods & BBBP & Tox21 & ToxCast & SIDER & ClinTox & MUV & HIV & BACE & PPI & A.R.$\downarrow$ \\
    \hline
    \hline
    No pre-train. & 65.8$\pm$4.5 & 74.0$\pm$0.8 & 63.4$\pm$0.6 & 57.3$\pm$1.6 & 58.0$\pm$4.4 & 71.8$\pm$2.5 & 75.3$\pm$1.9 & 70.1$\pm$5.4 & 64.8$\pm$1.0 & 6.6 \\
    \hline
    Infomax & 68.8$\pm$0.8 & 75.3$\pm$0.5 & 62.7$\pm$0.4 & 58.4$\pm$0.8 & 69.9$\pm$3.0 & \darkred{75.3}$\pm$2.5 & 76.0$\pm$0.7 & 75.9$\pm$1.6 & 64.1$\pm$1.5 & 5.3 \\
    EdgePred & 67.3$\pm$2.4 & \darkred{76.0}$\pm$0.6 & \darkred{64.1}$\pm$0.6 & 60.4$\pm$0.7 & 64.1$\pm$3.7 & 74.1$\pm$2.1 & 76.3$\pm$1.0 & \darkred{79.9}$\pm$0.9 & \darkred{65.7}$\pm$1.3 & 3.8 \\
    AttrMasking & 64.3$\pm$2.8 & \darkred{76.7}$\pm$0.4 & \darkred{64.2}$\pm$0.5 & \darkred{61.0}$\pm$0.7 & 71.8$\pm$4.1 & \darkred{74.7}$\pm$1.4 & 77.2$\pm$1.1 & \darkred{79.3}$\pm$1.6 & \darkred{65.2}$\pm$1.6 & 3.1 \\
    ContextPred & 68.0$\pm$2.0 & \darkred{75.7}$\pm$0.7 & \darkred{63.9}$\pm$0.6 & \darkred{60.9}$\pm$0.6 & 65.9$\pm$3.8 & \darkred{75.8}$\pm$1.7 & \darkred{77.3}$\pm$1.0 & \darkred{79.6}$\pm$1.2 & 64.4$\pm$1.3 & 3.4 \\
    GraphCL & \darkred{69.68}$\pm$0.67 & 73.87$\pm$0.66 & 62.40$\pm$0.57 & \darkred{60.53}$\pm$0.88 & \darkred{75.99}$\pm$2.65 & 69.80$\pm$2.66 & \darkred{78.47}$\pm$1.22 & 75.38$\pm$1.44 & \darkred{67.88}$\pm$0.85 & 4.6 \\
    \hline
    JOAO & \darkred{70.22}$\pm$0.98 & 74.98$\pm$0.29 & 62.94$\pm$0.48 & 59.97$\pm$0.79 & \darkred{81.32}$\pm$2.49 & 71.66$\pm$1.43 & 76.73$\pm$1.23 & 77.34$\pm$0.48 & 64.43$\pm$1.38 & 4.5 \\
    JOAOv2 & \darkred{71.39}$\pm$0.92 & 74.27$\pm$0.62 & 63.16$\pm$0.45 & 60.49$\pm$0.74 & \darkred{80.97}$\pm$1.64 & 73.67$\pm$1.00 & \darkred{77.51}$\pm$1.17 & 75.49$\pm$1.27 & 63.94$\pm$1.59 & 4.3 \\
    \hline
    \hline
\end{tabular}}
\end{center}
\end{table*}

\subsection{Results}
\subsubsection{On Diverse Datasets from TUDataset} \label{sec:exp_tudataset}
The results of semi-supervised learning \& unsupervised representation learning on TUDataset are in Tables \ref{tab:semi-supervised} \& \ref{tab:unsupervised}, respectively.
Through comparisons between (1) JOAO and GraphCL, (2) JOAOv2 and JOAO, and (3) JOAOv2 and heuristic self-supervised methods, we have the following observations.

\textbf{(i) With automated selection, JOAO is comparable to GraphCL with ad hoc rules from exhaustive manual tuning.}
With the automatic, adaptive, and dynamic augmentation selection procedure, JOAO performs  comparably to GraphCL whose augmentations are based on empirical ad-hoc rules gained from expensive trial-and-errors on the same TUDatase.
Specifically in semi-supervised learning (Table \ref{tab:semi-supervised}), JOAO 
matches or beats GraphCL in 7 out of 10 experiments, albeit with a slightly worse average rank.  
Similar observations were made in unsupervised learning (Table \ref{tab:unsupervised}).  
These results echo our earlier results 
in Section \ref{sec:joa_correlation} that automated selections of augmentation pairs made by JOAO were generally consistent with GraphCL's ``best practices" observed from manual tuning.   

\textbf{(ii) Augmentation-aware projection heads provide further improvement upon JOAO.}
With augmentation-aware projection heads introduced into JOAO, JOAOv2 further improves the performance and sometimes even outperforms GraphCL with the pre-defined rules of thumb for augmentation selections.
In semi-supervised learning (Table \ref{tab:semi-supervised}), JOAOv2 achieves the best average ranks of 2.0 and 2.8 under 1\% and 10\% label rate, respectively,
and in unsupervised representation learning (Table \ref{tab:unsupervised}) its average rank (2.8) is only edged by GraphCL (2.6).
The performance acquired by JOAOv2 echoes our conjecture in sec.~\ref{sec:joap} that such explicit disentanglement of the distorted feature distributions caused by various augmentation pairs would reel in the benefits from stronger augmentations.


\textbf{(iii) Across diverse datasets, JOAOv2 generally outperform heuristic self-supervised methods.}
On datasets originated from diverse sources, JOAOv2 generally outperforms heuristic self-supervised methods. Specifically in Table \ref{tab:semi-supervised}, JOAOv2 achieves no less than 0.3 average ranking gap with all heuristic self-supervised methods under 1\% label rate, and 1.0 with all but Infomax under 10\% label rate, which outperforms JOAO but still underperforms JOAOv2,
and in Table \ref{tab:unsupervised} only InfoGraph outperforms JOAO but underperforms JOAOv2 where there is no less that 1.5 average ranking gap between others and JOAOv2.
This in general meets our expectation that heuristic self-supervised methods can work well when guided by useful domain knowledge, which is hard to guarantee across diverse datasets.  In contrast JOAOv2 can dynamically and automatically adapt augmentation selections during self-supervised training, exploiting the signals (knowledge) from data.

\subsubsection{On Specific Bioinformatics Datasets.} \label{sec:transfer}
The results of transfer learning on bioinfomatics datasets are in Table \ref{tab:transfer}.
Similarly, through comparisons among JOAO(v2), GraphCL and heuristic self-supervised methods, we make the following findings.

\textbf{(iv) Without domain expertise incorporated, JOAOv2 underperforms some heuristic self-supervised methods in specific domains.}
Nevertheless, converse to that on diverse datasets, on the specific bioinformatics datasets, JOAOv2 underperforms some heuristic self-supervised methods designed with dataset-specific domain knowledge \cite{hu2019strategies} even though it improves average rank against GraphCL and JOAO.
As stated in Section \ref{sec:compared_algorithms}, the specific domain knowledge encoded in the compared heuristically designed methods correlates with the downstream datasets as shown in \cite{hu2019strategies}, which is not made available to benefit JOAOv2 that works well for a \textit{general} dataset (as shown in Section \ref{sec:exp_tudataset}), which may not suffice to capture the sophisticated domain expertise.  
Therefore, to make JOAOv2 even more competitive, domain knowledge can be introduced into the framework, for instance through proposing dataset-specific augmentation types and/or priors.
We leave this to future work.

\textbf{(v) With better generalizability, JOAOv2 outperforms GraphCL on unseen datasets.}
Different from results on diverse datasets from TUDataset, both JOAO and JOAOv2 outperform GraphCL with empirically pre-defined rules for augmentation selection on the unseen bioinfomatics datasets. Specifically in Table \ref{tab:transfer} JOAO improves the average rank by 0.1 and JOAOv2 did by 0.3 compared to GraphCL.
Note that the sampling rules of GraphCL were empirically derived from TUDataset hence these rules are not necessarily effective for the previously-unseen bioinfomatics datasets. In contrast, JOAOv2 dynamically and automatically learns the sampling distributions during self-supervised training, possessing the better generalizability. 
\begin{table}[t]
 \caption{Semi-supervised learning on large-scale OGB datasets.
 \darkred{Red} numbers indicate the top-2 performances (accuracy in \% on ogbg-ppa, F1 score in \% on ogbg-code).}
 \label{tab:ogb}
 \centering
 \resizebox{0.38\textwidth}{!}{
 \begin{tabular}{c | c | c c } 
  \hline
  \hline
  L.R. & Methods & ogbg-ppa & ogbg-code \\
  \hline
  \hline
  1\% & No pre-train. & 16.04$\pm$0.74 & 6.06$\pm$0.01 \\
  & GraphCL & 40.81$\pm$1.33 & \darkred{7.66}$\pm$0.25 \\
  \cdashline{2-4}
  & JOAO & \darkred{47.19}$\pm$1.30 & 6.84$\pm$0.31 \\
  & JOAOv2 & \darkred{44.30}$\pm$1.67 & \darkred{7.74}$\pm$0.24 \\
  \hline
  10\% & No pre-train. & 56.01$\pm$1.05 & 17.85$\pm$0.60 \\
  & GraphCL & 57.77$\pm$1.25 & \darkred{22.45}$\pm$0.17 \\
  \cdashline{2-4}
  & JOAO & \darkred{60.91}$\pm$0.83 & 22.06$\pm$0.30 \\
  & JOAOv2 & \darkred{59.32}$\pm$1.11 & \darkred{22.65}$\pm$0.22 \\
  \hline
  \hline
 \end{tabular}}
\end{table}

\subsubsection{On Large-Scale OGB Datasets.}
\textbf{(vi) JOAOv2 scales up well for large datasets.}
Both JOAO and JOAOv2 scale up for large datasets at least as well as GraphCL does. In Table \ref{tab:ogb}, for the ogbg-ppa dataset, JOAO improved even more significantly compared to GraphCL (by reference to earlier, smaller datasets), with $>$3.49\% and $>$1.55\% accuracy gains at 1\% and 10\% label rates, respectively.

\subsection{Summary of Main Findings}
We briefly summarize the aforementioned results as follows:\vspace{-1em}
\begin{itemize}
    \item Across datasets originated from diverse sources, JOAO with adaptive augmentation selection performs comparably to GraphCL, a strong baseline with exhaustively tuned augmentation rules by hand.\vspace{-0.5em}
    \item With augmentation-aware projection heads, JOAOv2 further boosts the performance and sometimes even outperforms GraphCL.\vspace{-0.5em}
\item On datasets from specific bioinformatics domains, JOAOv2 achieves better performance than  GraphCL whose empirical rules were not derived from such data, indicating its better generalizability to unseen datasets.\vspace{-0.5em}
\item Both JOAO and JOAOv2 outperform heuristic self-supervised methods with few exceptions. They might be further enhanced by encoding domain knowledge.\vspace{-0.5em}
\item JOAOv2 scales up to large datasets as well as GraphCL does, sometimes with even more significant improvement compared with that for smaller datasets.\vspace{-0.5em}
\end{itemize}

\section{Conclusions \& Discussions}
In this paper, we propose a unified bi-level optimization framework to dynamically and automatically select augmentations in GraphCL, named JOint Augmentation Optimization (JOAO).
The general framework is instantiated as min-max optimization, with empirical analysis showing that JOAO makes augmentation selections in general accordance with previous ``best practices" from exhaustive hand tuning for every dataset.
Furthermore, a new augmentation-aware projection head mechanism is proposed to overcome the potential training distribution distortion, resulting from the more aggressive and varying augmentations by JOAO. Experiments demonstrate that JOAO and its variant performs on par with and sometimes better than the state-of-the-art competitors including GraphCL on multiple graph datasets of various scales and types, yet without resorting to tedious dataset-specific manual tuning.

Although JOAO automates GraphCL in selecting augmentation pairs, it still relies on human prior knowledge in constructing and configuring the augmentation  pool to select from.  In this sense ``full'' automation is still desired and will be pursued in future work.
Meanwhile, in parallel to the principled formulation of bi-level optimization, a meta-learning formulation can also be pursued.




\bibliography{example_paper}

\begin{thebibliography}{69}
\providecommand{\natexlab}[1]{#1}
\providecommand{\url}[1]{\texttt{#1}}
\expandafter\ifx\csname urlstyle\endcsname\relax
  \providecommand{\doi}[1]{doi: #1}\else
  \providecommand{\doi}{doi: \begingroup \urlstyle{rm}\Url}\fi

\bibitem[Adhikari et~al.(2018)Adhikari, Zhang, Ramakrishnan, and
  Prakash]{adhikari2018sub2vec}
Adhikari, B., Zhang, Y., Ramakrishnan, N., and Prakash, B.~A.
\newblock Sub2vec: Feature learning for subgraphs.
\newblock In \emph{Pacific-Asia Conference on Knowledge Discovery and Data
  Mining}, pp.\  170--182. Springer, 2018.

\bibitem[Baydin et~al.(2017)Baydin, Cornish, Rubio, Schmidt, and
  Wood]{baydin2017online}
Baydin, A.~G., Cornish, R., Rubio, D.~M., Schmidt, M., and Wood, F.
\newblock Online learning rate adaptation with hypergradient descent.
\newblock \emph{arXiv preprint arXiv:1703.04782}, 2017.

\bibitem[Boyd et~al.(2004)Boyd, Boyd, and Vandenberghe]{boyd2004convex}
Boyd, S., Boyd, S.~P., and Vandenberghe, L.
\newblock \emph{Convex optimization}.
\newblock Cambridge university press, 2004.

\bibitem[Chen et~al.(2020{\natexlab{a}})Chen, Zhang, Zhang, Tang, Cai, Chen,
  Li, Zhang, and Tang]{chen2020coad}
Chen, B., Zhang, J., Zhang, X., Tang, X., Cai, L., Chen, H., Li, C., Zhang, P.,
  and Tang, J.
\newblock {COAD}: Contrastive pre-training with adversarial fine-tuning for
  zero-shot expert linking.
\newblock \emph{arXiv preprint arXiv:2012.11336}, 2020{\natexlab{a}}.

\bibitem[Chen et~al.(2020{\natexlab{b}})Chen, Lin, Li, Li, Zhou, Sun,
  et~al.]{chen2020distance}
Chen, D., Lin, Y., Li, L., Li, X.~R., Zhou, J., Sun, X., et~al.
\newblock Distance-wise graph contrastive learning.
\newblock \emph{arXiv preprint arXiv:2012.07437}, 2020{\natexlab{b}}.

\bibitem[Chen et~al.(2019)Chen, Bian, and Sun]{chen2019powerful}
Chen, T., Bian, S., and Sun, Y.
\newblock Are powerful graph neural nets necessary? a dissection on graph
  classification.
\newblock \emph{arXiv preprint arXiv:1905.04579}, 2019.

\bibitem[Chen et~al.(2020{\natexlab{c}})Chen, Kornblith, Norouzi, and
  Hinton]{chen2020simple}
Chen, T., Kornblith, S., Norouzi, M., and Hinton, G.
\newblock A simple framework for contrastive learning of visual
  representations.
\newblock In \emph{International conference on machine learning}, pp.\
  1597--1607. PMLR, 2020{\natexlab{c}}.

\bibitem[Dobson \& Doig(2003)Dobson and Doig]{dobson2003distinguishing}
Dobson, P.~D. and Doig, A.~J.
\newblock Distinguishing enzyme structures from non-enzymes without alignments.
\newblock \emph{Journal of molecular biology}, 330\penalty0 (4):\penalty0
  771--783, 2003.

\bibitem[Dwivedi et~al.(2020)Dwivedi, Joshi, Laurent, Bengio, and
  Bresson]{dwivedi2020benchmarking}
Dwivedi, V.~P., Joshi, C.~K., Laurent, T., Bengio, Y., and Bresson, X.
\newblock Benchmarking graph neural networks.
\newblock \emph{arXiv preprint arXiv:2003.00982}, 2020.

\bibitem[Franceschi et~al.(2017)Franceschi, Donini, Frasconi, and
  Pontil]{franceschi2017forward}
Franceschi, L., Donini, M., Frasconi, P., and Pontil, M.
\newblock Forward and reverse gradient-based hyperparameter optimization.
\newblock In \emph{International Conference on Machine Learning}, pp.\
  1165--1173. PMLR, 2017.

\bibitem[Gould et~al.(2016)Gould, Fernando, Cherian, Anderson, Cruz, and
  Guo]{gould2016differentiating}
Gould, S., Fernando, B., Cherian, A., Anderson, P., Cruz, R.~S., and Guo, E.
\newblock On differentiating parameterized argmin and argmax problems with
  application to bi-level optimization.
\newblock \emph{arXiv preprint arXiv:1607.05447}, 2016.

\bibitem[Grover \& Leskovec(2016)Grover and Leskovec]{grover2016node2vec}
Grover, A. and Leskovec, J.
\newblock node2vec: Scalable feature learning for networks.
\newblock In \emph{Proceedings of the 22nd ACM SIGKDD international conference
  on Knowledge discovery and data mining}, pp.\  855--864, 2016.

\bibitem[Guiasu \& Shenitzer(1985)Guiasu and Shenitzer]{guiasu1985principle}
Guiasu, S. and Shenitzer, A.
\newblock The principle of maximum entropy.
\newblock \emph{The mathematical intelligencer}, 7\penalty0 (1):\penalty0
  42--48, 1985.

\bibitem[Hassani \& Khasahmadi(2020)Hassani and
  Khasahmadi]{hassani2020contrastive}
Hassani, K. and Khasahmadi, A.~H.
\newblock Contrastive multi-view representation learning on graphs.
\newblock \emph{arXiv preprint arXiv:2006.05582}, 2020.

\bibitem[He et~al.(2020)He, Fan, Wu, Xie, and Girshick]{he2020momentum}
He, K., Fan, H., Wu, Y., Xie, S., and Girshick, R.
\newblock Momentum contrast for unsupervised visual representation learning.
\newblock In \emph{Proceedings of the IEEE/CVF Conference on Computer Vision
  and Pattern Recognition}, 2020.

\bibitem[Hu et~al.(2019)Hu, Liu, Gomes, Zitnik, Liang, Pande, and
  Leskovec]{hu2019strategies}
Hu, W., Liu, B., Gomes, J., Zitnik, M., Liang, P., Pande, V., and Leskovec, J.
\newblock Strategies for pre-training graph neural networks.
\newblock \emph{arXiv preprint arXiv:1905.12265}, 2019.

\bibitem[Hu et~al.(2020{\natexlab{a}})Hu, Fey, Zitnik, Dong, Ren, Liu, Catasta,
  and Leskovec]{hu2020open}
Hu, W., Fey, M., Zitnik, M., Dong, Y., Ren, H., Liu, B., Catasta, M., and
  Leskovec, J.
\newblock Open graph benchmark: Datasets for machine learning on graphs.
\newblock \emph{arXiv preprint arXiv:2005.00687}, 2020{\natexlab{a}}.

\bibitem[Hu et~al.(2020{\natexlab{b}})Hu, Dong, Wang, Chang, and
  Sun]{hu2020gpt}
Hu, Z., Dong, Y., Wang, K., Chang, K.-W., and Sun, Y.
\newblock {GPT-GNN}: Generative pre-training of graph neural networks.
\newblock In \emph{Proceedings of the 26th ACM SIGKDD International Conference
  on Knowledge Discovery \& Data Mining}, pp.\  1857--1867, 2020{\natexlab{b}}.

\bibitem[Huang et~al.(2021)Huang, Pei, Menkovski, and
  Pechenizkiy]{huang2021hop}
Huang, T., Pei, Y., Menkovski, V., and Pechenizkiy, M.
\newblock Hop-count based self-supervised anomaly detection on attributed
  networks.
\newblock \emph{arXiv preprint arXiv:2104.07917}, 2021.

\bibitem[Hwang et~al.(2020)Hwang, Park, Kwon, Kim, Ha, and Kim]{hwang2020self}
Hwang, D., Park, J., Kwon, S., Kim, K., Ha, J.-W., and Kim, H.~J.
\newblock Self-supervised auxiliary learning with meta-paths for heterogeneous
  graphs.
\newblock \emph{Advances in Neural Information Processing Systems}, 33, 2020.

\bibitem[Jin et~al.(2021{\natexlab{a}})Jin, Zheng, Li, Gong, Zhou, and
  Pan]{jin2021multi}
Jin, M., Zheng, Y., Li, Y.-F., Gong, C., Zhou, C., and Pan, S.
\newblock Multi-scale contrastive siamese networks for self-supervised graph
  representation learning.
\newblock \emph{arXiv preprint arXiv:2105.05682}, 2021{\natexlab{a}}.

\bibitem[Jin et~al.(2020)Jin, Derr, Liu, Wang, Wang, Liu, and
  Tang]{jin2020self}
Jin, W., Derr, T., Liu, H., Wang, Y., Wang, S., Liu, Z., and Tang, J.
\newblock Self-supervised learning on graphs: Deep insights and new direction.
\newblock \emph{arXiv preprint arXiv:2006.10141}, 2020.

\bibitem[Jin et~al.(2021{\natexlab{b}})Jin, Liu, Zhao, Ma, Shah, and
  Tang]{jin2021automated}
Jin, W., Liu, X., Zhao, X., Ma, Y., Shah, N., and Tang, J.
\newblock Automated self-supervised learning for graphs, 2021{\natexlab{b}}.

\bibitem[Jun et~al.(2020)Jun, Child, Chen, Schulman, Ramesh, Radford, and
  Sutskever]{jun2020distribution}
Jun, H., Child, R., Chen, M., Schulman, J., Ramesh, A., Radford, A., and
  Sutskever, I.
\newblock Distribution augmentation for generative modeling.
\newblock In \emph{International Conference on Machine Learning}, pp.\
  5006--5019. PMLR, 2020.

\bibitem[Kipf \& Welling(2016)Kipf and Welling]{kipf2016variational}
Kipf, T.~N. and Welling, M.
\newblock Variational graph auto-encoders.
\newblock \emph{arXiv preprint arXiv:1611.07308}, 2016.

\bibitem[Kong et~al.(2020)Kong, Li, Ding, Wu, Zhu, Ghanem, Taylor, and
  Goldstein]{kong2020flag}
Kong, K., Li, G., Ding, M., Wu, Z., Zhu, C., Ghanem, B., Taylor, G., and
  Goldstein, T.
\newblock {FLAG}: Adversarial data augmentation for graph neural networks.
\newblock \emph{arXiv preprint arXiv:2010.09891}, 2020.

\bibitem[Lee et~al.(2020)Lee, Hwang, and Shin]{lee2020self}
Lee, H., Hwang, S.~J., and Shin, J.
\newblock Self-supervised label augmentation via input transformations.
\newblock In \emph{International Conference on Machine Learning}, pp.\
  5714--5724. PMLR, 2020.

\bibitem[Li et~al.(2021)Li, Huang, and Zitnik]{li2021representation}
Li, M.~M., Huang, K., and Zitnik, M.
\newblock Representation learning for networks in biology and medicine:
  Advancements, challenges, and opportunities.
\newblock \emph{arXiv preprint arXiv:2104.04883}, 2021.

\bibitem[Liu et~al.(2020)Liu, Gao, and Ji]{liu2020towards}
Liu, M., Gao, H., and Ji, S.
\newblock Towards deeper graph neural networks.
\newblock In \emph{Proceedings of the 26th ACM SIGKDD International Conference
  on Knowledge Discovery \& Data Mining}, pp.\  338--348, 2020.

\bibitem[Liu et~al.(2021)Liu, Pan, Jin, Zhou, Xia, and Yu]{liu2021graph}
Liu, Y., Pan, S., Jin, M., Zhou, C., Xia, F., and Yu, P.~S.
\newblock Graph self-supervised learning: A survey.
\newblock \emph{arXiv preprint arXiv:2103.00111}, 2021.

\bibitem[Luketina et~al.(2016)Luketina, Berglund, Greff, and
  Raiko]{luketina2016scalable}
Luketina, J., Berglund, M., Greff, K., and Raiko, T.
\newblock Scalable gradient-based tuning of continuous regularization
  hyperparameters.
\newblock In \emph{International conference on machine learning}, pp.\
  2952--2960. PMLR, 2016.

\bibitem[Maclaurin et~al.(2015)Maclaurin, Duvenaud, and
  Adams]{maclaurin2015gradient}
Maclaurin, D., Duvenaud, D., and Adams, R.
\newblock Gradient-based hyperparameter optimization through reversible
  learning.
\newblock In \emph{International conference on machine learning}, pp.\
  2113--2122. PMLR, 2015.

\bibitem[Manessi \& Rozza(2020)Manessi and Rozza]{manessi2020graph}
Manessi, F. and Rozza, A.
\newblock Graph-based neural network models with multiple self-supervised
  auxiliary tasks.
\newblock \emph{arXiv preprint arXiv:2011.07267}, 2020.

\bibitem[Morris et~al.(2020)Morris, Kriege, Bause, Kersting, Mutzel, and
  Neumann]{morris2020tudataset}
Morris, C., Kriege, N.~M., Bause, F., Kersting, K., Mutzel, P., and Neumann, M.
\newblock Tudataset: A collection of benchmark datasets for learning with
  graphs.
\newblock \emph{arXiv preprint arXiv:2007.08663}, 2020.

\bibitem[Narayanan et~al.(2017)Narayanan, Chandramohan, Venkatesan, Chen, Liu,
  and Jaiswal]{narayanan2017graph2vec}
Narayanan, A., Chandramohan, M., Venkatesan, R., Chen, L., Liu, Y., and
  Jaiswal, S.
\newblock graph2vec: Learning distributed representations of graphs.
\newblock \emph{arXiv preprint arXiv:1707.05005}, 2017.

\bibitem[Nene et~al.(1996)Nene, Nayar, Murase, et~al.]{nene1996columbia}
Nene, S.~A., Nayar, S.~K., Murase, H., et~al.
\newblock Columbia object image library (coil-100).
\newblock 1996.

\bibitem[Park et~al.(2020)Park, Kim, Han, and Yu]{park2020unsupervised}
Park, C., Kim, D., Han, J., and Yu, H.
\newblock Unsupervised attributed multiplex network embedding.
\newblock In \emph{AAAI}, pp.\  5371--5378, 2020.

\bibitem[Pedregosa(2016)]{pedregosa2016hyperparameter}
Pedregosa, F.
\newblock Hyperparameter optimization with approximate gradient.
\newblock In \emph{International conference on machine learning}, pp.\
  737--746. PMLR, 2016.

\bibitem[Peng et~al.(2020{\natexlab{a}})Peng, Dong, Luo, Wu, and
  Zheng]{peng2020self}
Peng, Z., Dong, Y., Luo, M., Wu, X.-M., and Zheng, Q.
\newblock Self-supervised graph representation learning via global context
  prediction.
\newblock \emph{arXiv:2003.01604}, 2020{\natexlab{a}}.

\bibitem[Peng et~al.(2020{\natexlab{b}})Peng, Huang, Luo, Zheng, Rong, Xu, and
  Huang]{peng2020graph}
Peng, Z., Huang, W., Luo, M., Zheng, Q., Rong, Y., Xu, T., and Huang, J.
\newblock Graph representation learning via graphical mutual information
  maximization.
\newblock In \emph{Proceedings of The Web Conference 2020}, pp.\  259--270,
  2020{\natexlab{b}}.

\bibitem[Qiu et~al.(2020)Qiu, Chen, Dong, Zhang, Yang, Ding, Wang, and
  Tang]{qiu2020gcc}
Qiu, J., Chen, Q., Dong, Y., Zhang, J., Yang, H., Ding, M., Wang, K., and Tang,
  J.
\newblock {GCC}: Graph contrastive coding for graph neural network
  pre-training.
\newblock In \emph{Proceedings of the 26th ACM SIGKDD International Conference
  on Knowledge Discovery \& Data Mining}, 2020.

\bibitem[Ren et~al.(2019)Ren, Liu, Huang, Dai, Bo, and
  Zhang]{ren2019heterogeneous}
Ren, Y., Liu, B., Huang, C., Dai, P., Bo, L., and Zhang, J.
\newblock Heterogeneous deep graph infomax.
\newblock \emph{arXiv preprint arXiv:1911.08538}, 2019.

\bibitem[Riesen \& Bunke(2008)Riesen and Bunke]{riesen2008iam}
Riesen, K. and Bunke, H.
\newblock Iam graph database repository for graph based pattern recognition and
  machine learning.
\newblock In \emph{Joint IAPR International Workshops on Statistical Techniques
  in Pattern Recognition (SPR) and Structural and Syntactic Pattern Recognition
  (SSPR)}, pp.\  287--297. Springer, 2008.

\bibitem[Robey et~al.(2020)Robey, Hassani, and Pappas]{robey2020model}
Robey, A., Hassani, H., and Pappas, G.~J.
\newblock Model-based robust deep learning.
\newblock \emph{arXiv preprint arXiv:2005.10247}, 2020.

\bibitem[Rong et~al.(2020)Rong, Bian, Xu, Xie, Wei, Huang, and
  Huang]{rong2020self}
Rong, Y., Bian, Y., Xu, T., Xie, W., Wei, Y., Huang, W., and Huang, J.
\newblock Self-supervised graph transformer on large-scale molecular data.
\newblock \emph{Advances in Neural Information Processing Systems}, 33, 2020.

\bibitem[Roy et~al.(2021)Roy, Roy, Rahman, Amin, and Ali]{roy2021node}
Roy, K.~K., Roy, A., Rahman, A., Amin, M.~A., and Ali, A.~A.
\newblock Node embedding using mutual information and self-supervision based
  bi-level aggregation.
\newblock \emph{arXiv preprint arXiv:2104.13014}, 2021.

\bibitem[Rozemberczki et~al.(2020)Rozemberczki, Kiss, and
  Sarkar]{rozemberczki2020api}
Rozemberczki, B., Kiss, O., and Sarkar, R.
\newblock An {API} oriented open-source python framework for unsupervised
  learning on graphs.
\newblock \emph{arXiv preprint arXiv:2003.04819}, 2020.

\bibitem[Shaban et~al.(2019)Shaban, Cheng, Hatch, and
  Boots]{shaban2019truncated}
Shaban, A., Cheng, C.-A., Hatch, N., and Boots, B.
\newblock Truncated back-propagation for bilevel optimization.
\newblock In \emph{The 22nd International Conference on Artificial Intelligence
  and Statistics}, pp.\  1723--1732. PMLR, 2019.

\bibitem[Shervashidze et~al.(2009)Shervashidze, Vishwanathan, Petri, Mehlhorn,
  and Borgwardt]{shervashidze2009efficient}
Shervashidze, N., Vishwanathan, S., Petri, T., Mehlhorn, K., and Borgwardt, K.
\newblock Efficient graphlet kernels for large graph comparison.
\newblock In \emph{Artificial intelligence and statistics}, pp.\  488--495.
  PMLR, 2009.

\bibitem[Shervashidze et~al.(2011)Shervashidze, Schweitzer, Van~Leeuwen,
  Mehlhorn, and Borgwardt]{shervashidze2011weisfeiler}
Shervashidze, N., Schweitzer, P., Van~Leeuwen, E.~J., Mehlhorn, K., and
  Borgwardt, K.~M.
\newblock Weisfeiler-{L}ehman graph kernels.
\newblock \emph{Journal of Machine Learning Research}, 12\penalty0 (9), 2011.

\bibitem[Snoek et~al.(2012)Snoek, Larochelle, and Adams]{snoek2012practical}
Snoek, J., Larochelle, H., and Adams, R.~P.
\newblock Practical bayesian optimization of machine learning algorithms.
\newblock \emph{Advances in Neural Information Processing Systems}, 2012.

\bibitem[Srinivas et~al.(2010)Srinivas, Krause, Kakade, and
  Seeger]{srinivas2010gaussian}
Srinivas, N., Krause, A., Kakade, S., and Seeger, M.
\newblock Gaussian process optimization in the bandit setting: No regret and
  experimental design.
\newblock In \emph{Proceedings of the International Conference on Machine
  Learning, 2010}, 2010.

\bibitem[Sun et~al.(2019)Sun, Hoffmann, Verma, and Tang]{sun2019infograph}
Sun, F.-Y., Hoffmann, J., Verma, V., and Tang, J.
\newblock Info{G}raph: Unsupervised and semi-supervised graph-level
  representation learning via mutual information maximization.
\newblock \emph{arXiv preprint arXiv:1908.01000}, 2019.

\bibitem[Veli{\v{c}}kovi{\'c} et~al.(2018)Veli{\v{c}}kovi{\'c}, Fedus,
  Hamilton, Li{\`o}, Bengio, and Hjelm]{velivckovic2018deep}
Veli{\v{c}}kovi{\'c}, P., Fedus, W., Hamilton, W.~L., Li{\`o}, P., Bengio, Y.,
  and Hjelm, R.~D.
\newblock Deep graph infomax.
\newblock \emph{arXiv preprint arXiv:1809.10341}, 2018.

\bibitem[Wang \& Liu(2021)Wang and Liu]{wang2021learning}
Wang, C. and Liu, Z.
\newblock Learning graph representation by aggregating subgraphs via mutual
  information maximization.
\newblock \emph{arXiv preprint arXiv:2103.13125}, 2021.

\bibitem[Wang et~al.(2019)Wang, Zhang, Liu, Chen, Xu, Fardad, and
  Li]{wang2019towards}
Wang, J., Zhang, T., Liu, S., Chen, P.-Y., Xu, J., Fardad, M., and Li, B.
\newblock Towards a unified min-max framework for adversarial exploration and
  robustness.
\newblock \emph{arXiv preprint arXiv:1906.03563}, 2019.

\bibitem[Xie et~al.(2020)Xie, Tan, Gong, Wang, Yuille, and
  Le]{xie2020adversarial}
Xie, C., Tan, M., Gong, B., Wang, J., Yuille, A.~L., and Le, Q.~V.
\newblock Adversarial examples improve image recognition.
\newblock In \emph{Proceedings of the IEEE/CVF Conference on Computer Vision
  and Pattern Recognition}, 2020.

\bibitem[Xie et~al.(2021)Xie, Xu, Zhang, Wang, and Ji]{xie2021self}
Xie, Y., Xu, Z., Zhang, J., Wang, Z., and Ji, S.
\newblock Self-supervised learning of graph neural networks: A unified review.
\newblock \emph{arXiv preprint arXiv:2102.10757}, 2021.

\bibitem[Xu et~al.(2018)Xu, Hu, Leskovec, and Jegelka]{xu2018powerful}
Xu, K., Hu, W., Leskovec, J., and Jegelka, S.
\newblock How powerful are graph neural networks?
\newblock \emph{arXiv preprint arXiv:1810.00826}, 2018.

\bibitem[Yanardag \& Vishwanathan(2015)Yanardag and
  Vishwanathan]{yanardag2015deep}
Yanardag, P. and Vishwanathan, S.
\newblock Deep graph kernels.
\newblock In \emph{Proceedings of the 21th ACM SIGKDD international conference
  on knowledge discovery and data mining}, pp.\  1365--1374, 2015.

\bibitem[You \& Shen(2020)You and Shen]{you2020cross}
You, Y. and Shen, Y.
\newblock Cross-modality protein embedding for compound-protein affinity and
  contact prediction.
\newblock \emph{arXiv preprint arXiv:2012.00651}, 2020.

\bibitem[You et~al.(2020{\natexlab{a}})You, Chen, Sui, Chen, Wang, and
  Shen]{you2020graph}
You, Y., Chen, T., Sui, Y., Chen, T., Wang, Z., and Shen, Y.
\newblock Graph contrastive learning with augmentations.
\newblock \emph{Advances in Neural Information Processing Systems}, 33,
  2020{\natexlab{a}}.

\bibitem[You et~al.(2020{\natexlab{b}})You, Chen, Wang, and Shen]{you2020does}
You, Y., Chen, T., Wang, Z., and Shen, Y.
\newblock When does self-supervision help graph convolutional networks?
\newblock In \emph{International Conference on Machine Learning}, pp.\
  10871--10880. PMLR, 2020{\natexlab{b}}.

\bibitem[You et~al.(2020{\natexlab{c}})You, Chen, Wang, and Shen]{you2020l2}
You, Y., Chen, T., Wang, Z., and Shen, Y.
\newblock L$^2$-{GCN}: Layer-wise and learned efficient training of graph
  convolutional networks.
\newblock In \emph{Proceedings of the IEEE/CVF Conference on Computer Vision
  and Pattern Recognition}, pp.\  2127--2135, 2020{\natexlab{c}}.

\bibitem[Zhang et~al.(2020)Zhang, Lin, Liu, Zhou, Tang, Liang, and
  Xing]{zhang2020iterative}
Zhang, H., Lin, S., Liu, W., Zhou, P., Tang, J., Liang, X., and Xing, E.~P.
\newblock Iterative graph self-distillation.
\newblock \emph{arXiv preprint arXiv:2010.12609}, 2020.

\bibitem[Zhao et~al.(2020)Zhao, Liu, Neves, Woodford, Jiang, and
  Shah]{zhao2020data}
Zhao, T., Liu, Y., Neves, L., Woodford, O., Jiang, M., and Shah, N.
\newblock Data augmentation for graph neural networks.
\newblock \emph{arXiv preprint arXiv:2006.06830}, 2020.

\bibitem[Zhu et~al.(2020{\natexlab{a}})Zhu, Du, and Yan]{zhu2020self}
Zhu, Q., Du, B., and Yan, P.
\newblock Self-supervised training of graph convolutional networks.
\newblock \emph{arXiv preprint arXiv:2006.02380}, 2020{\natexlab{a}}.

\bibitem[Zhu et~al.(2020{\natexlab{b}})Zhu, Xu, Yu, Liu, Wu, and
  Wang]{zhu2020deep}
Zhu, Y., Xu, Y., Yu, F., Liu, Q., Wu, S., and Wang, L.
\newblock Deep graph contrastive representation learning.
\newblock \emph{arXiv preprint arXiv:2006.04131}, 2020{\natexlab{b}}.

\bibitem[Zhu et~al.(2020{\natexlab{c}})Zhu, Xu, Yu, Liu, Wu, and
  Wang]{zhu2020graph}
Zhu, Y., Xu, Y., Yu, F., Liu, Q., Wu, S., and Wang, L.
\newblock Graph contrastive learning with adaptive augmentation.
\newblock \emph{arXiv preprint arXiv:2010.14945}, 2020{\natexlab{c}}.

\end{thebibliography}
\bibliographystyle{icml2021}

\end{document}